# Magnet-Based Soft Robotic Skin Using a 3D-Printed Multi-Lattice Structure and CNN-Based Tactile Super-Resolution

[†]Yunseong Bang, [†]Joowon Park, Suan Sim, Youngjun Ryu, Sukho Park, and [*]Kyungseo Park

*Abstract*— **This paper presents a magnet-based robotic skin that integrates a multilayer soft lattice with distributed Hall-effect sensor arrays and a tactile super-resolution model. External contact forces are converted to magnetic field changes by embedded permanent magnets, and the lattice spreads these changes across the sensing domain. This gives each sensor a large, overlapping receptive field and enables a large sensing area with minimal blind spots. Lattice parameters are tunable, enabling joint adjustment of mechanical compliance and transduction characteristics. An implicit modeling workflow and selective laser sintering (SLS) 3D printing support rapid fabrication of conformal, high-complexity structures. A convolutional neural network trained on experimental measurements estimates contact location and normal force in real time. Experiments validate localization accuracy and indicate scalability to larger surfaces, suggesting applicability to whole-body robotic skin and safe human-robot interaction.**

## I. INTRODUCTION

In human-robot coexistence settings, soft whole-body robotic skin should enable tactile perception and shock absorption to achieve safe physical human-robot interaction [1], [2], [3]. One of the most common ways to realize whole-body robotic skin is deploying discrete tactile sensor modules [4], [5]. This approach achieves a large sensing area efficiently, but it has the drawbacks that the sensing elements may be directly exposed to the environment. To address this, pneumatic robotic skins that physically isolate fragile sensors from external stimuli and employ impact-resistant TPU have been explored, but the low spatial resolution is considered a main limitation [6]. Vision-based robotic skin has also been proposed, but inherently require a sufficient field of view inside the robot, creating major design constraints [7].

Meanwhile, many researchers have explored AI-driven tactile super-resolution that achieves spatial resolution beyond the physical sensor layout[8]. This approach deliberately overlaps receptive fields of sensors, and reconstructs dense tactile information by interpolating across signals. A representative example is electrical impedance tomography (EIT)-based robotic skin [9]; in this approach, electrodes are distributed across a continuous piezoresistive domain, and impedance is measured via multiplexed electrode pairs to reconstruct the conductivity map.

Y. Bang, S. Sim, Y. Ryu, S. Park, and K. Park are with the Department of Robotics and Mechatronics Engineering, DGIST (Daegu Gyeongbuk Institute of Science and Technology), 42988 Daegu, Republic of Korea

J. Park is with Department of Electrical, Electronic and Computer Engineering, University of Ulsan, 93 Daehak-ro, Nam-gu, 44610 Ulsan, Republic of Korea

[†]These authors contributed equally to this work.

[*]Corresponding authors: kspark@dgist.ac.kr

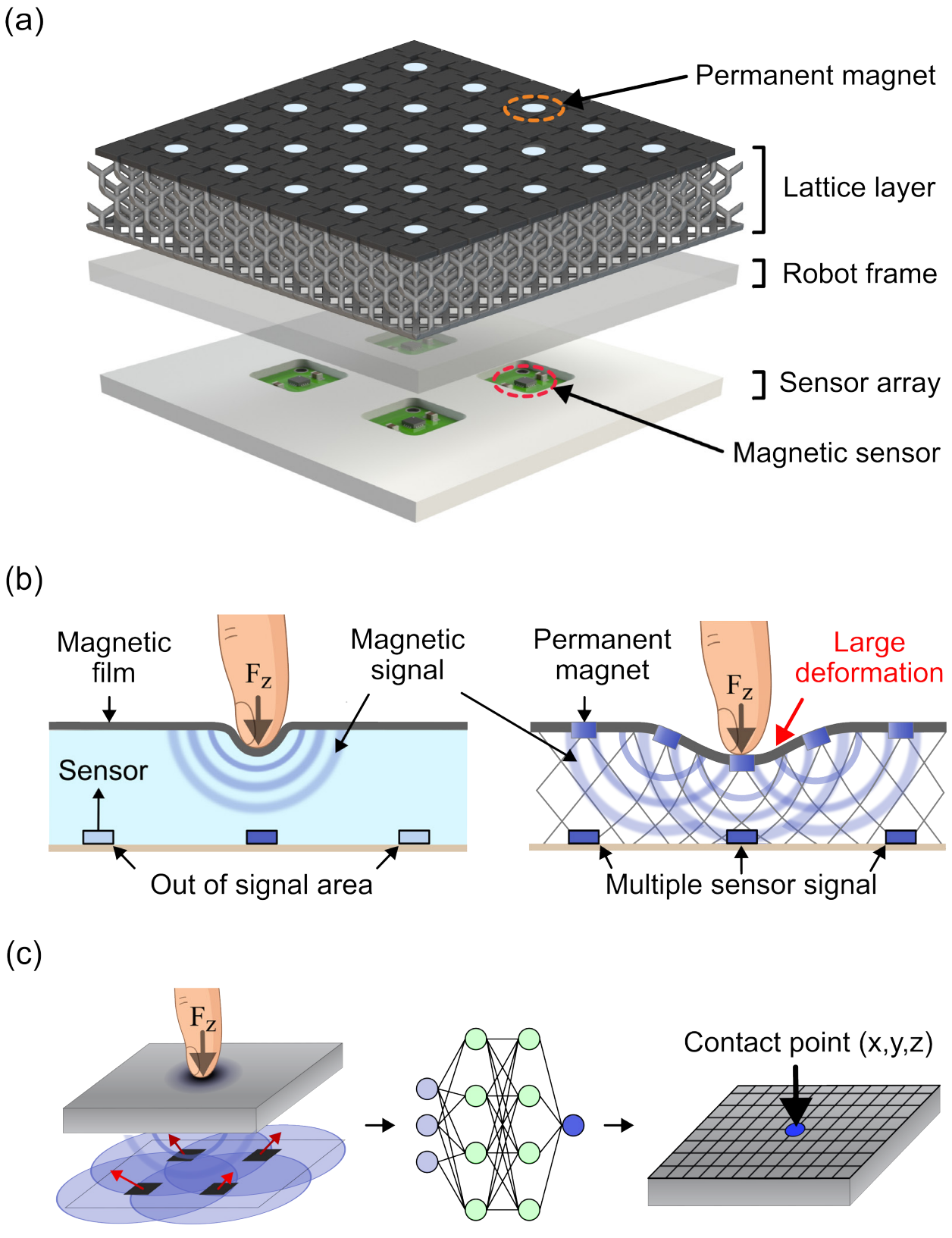

Fig. 1. Concept of proposed robotic skin. (a) Structure of proposed robotic skin. (b) Mechanism of conventional method and our proposed robotic skin (c) super-resolution tactile sensing

Using an elastomeric magnet with Hall-effect sensors is a noteworthy alternative for implementing tactile super-resolution. For instance, sinusoidal magnetization of the film achieves self-decoupled three-axis force sensing [10], [11], enabling three-axis force reconstruction. However, magnetized films produce much weaker fields than permanent magnets, and the field strength decreases as the elastomer becomes softer. Furthermore, typical silicone elastomers are highly compliant, so deformation under external loading can be strongly localized [8]. This means that force-induced changes in the magnetic field remain confined near the contact, and achieving large-area coverage requires densely packed sensors, limiting scalability.

In this paper, we propose a magnetic robotic skin that employs a 3D-printed lattice structure and evenly spaced

Hall-effect sensor arrays, as shown in Fig. 1(a). The soft lattice structure was made of TPU, and it was printed using a selective laser sintering (SLS) 3D printer. This design incorporates permanent magnets that convert external forces into variations in the magnetic field. The resulting perturbations spread around the contact point and are measured by an array of Hall-effect sensors. Using a convolutional neural network (CNN) trained on experimental data, we infer the stimulus location and magnitude, achieving real-time tactile super-resolution. The main contributions of this work are as follows.

(1) Development of a soft multi-lattice structure to create overlapping receptive fields.

(2) Fabrication of the magnet-embedded TPU lattice structure through SLS-type 3D printing.

(3) Implementation of a sensing architecture physically isolating Hall-effect sensors from external stimuli.

(4) Validation of an approach that infers the location and magnitude of physical contact from real-world data.

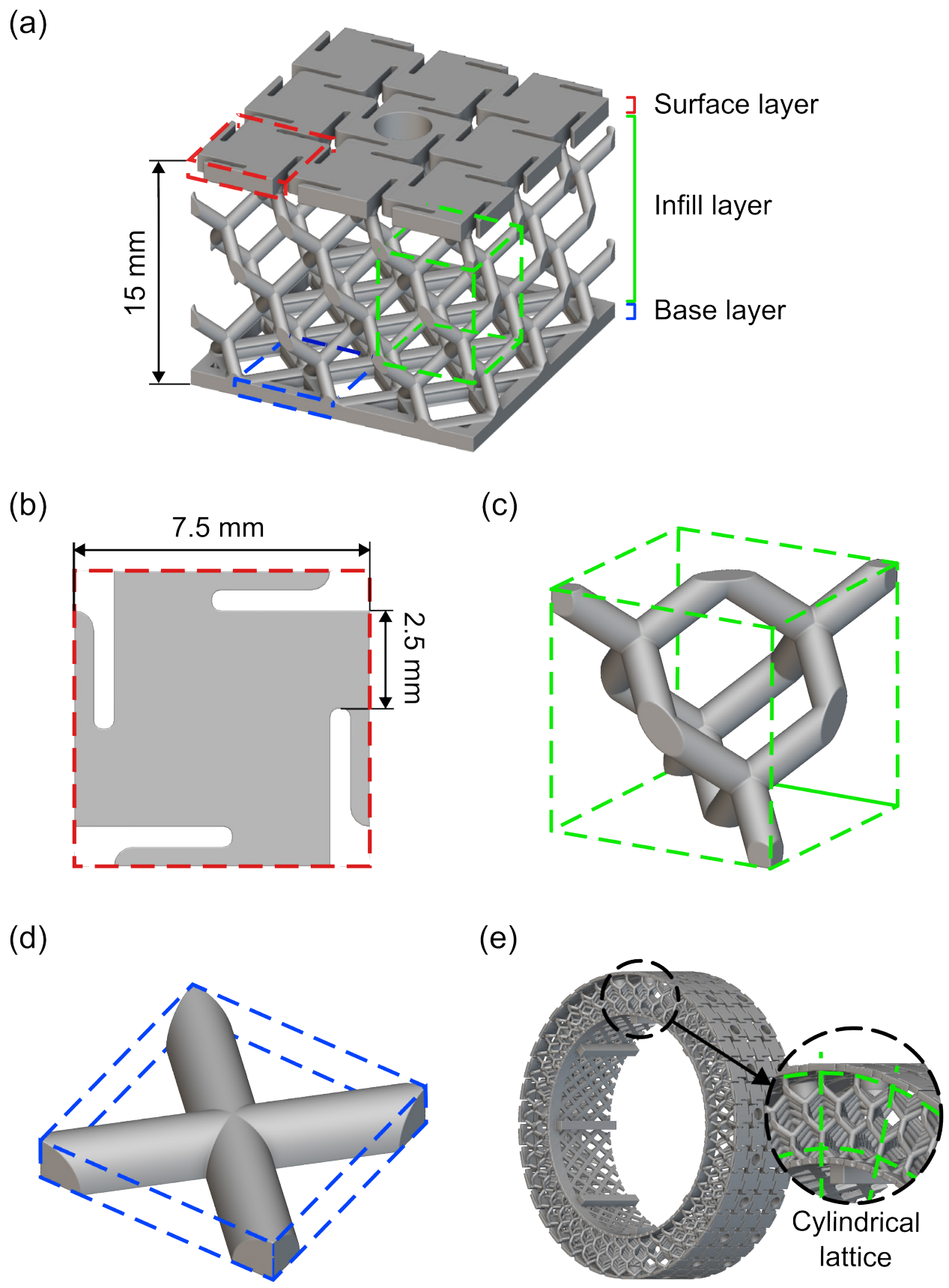

Fig. 2. Hardware Design: (a) lattice structure of proposed robotic skin, (b) surface unit cell, (c) 3D diamond unit cell, (d) 2D diamond unit cell, and (e) scalability on curved surface.

## II. HARDWARE DESIGN

Several approaches have been explored to overcome the scalability limitations of magnetized-film-based tactile skins. Recent work addressed this by embedding magnetic films into a large silicone substrate with a sparse Hall-effect sensor array [12]; however, this approach estimates only XY contact position and does not recover depth or normal force along the Z-axis. To overcome the scalability and practicality limitations of magnetized films, a whole-body magnetic robotic skin was realized by attaching permanent magnets to sponge-foam modules [13]. However, the design still suffers from blind spots due to physical gaps between the foam modules.

Here, we propose a compliant multi-lattice structure that effectively realizes overlapping receptive fields. The structure converts normal forces into distributed deformation, inducing translation and rotation of surface-mounted neodymium magnets. The resulting wide-area magnetic-field variations allow the Hall-effect sensor array to be deployed at a substantially lower spatial density.

### A. Lattice Structure

In this work, the proposed multi-lattice structure was designed using nTop, a design program primarily used in studies involving lattice type and parameter modifications [14], [15]. Unlike traditional CAD, which relies on boundary-representation features, this program employs implicit modeling that describes 3D geometry with continuous mathematical equations, making it more robust and better suited to complex parts such as lattices. Our lattice design comprises surface, infill, and base layers, and each part is generated from unit cells with distinct geometry. The unit-cell size was fixed at 7.5 mm × 7.5 mm across all layers.

The surface layer was designed to spread localized contact into the infill layer. Because TPU has limited in-plane stretchability, a solid, unpatterned surface tends to wrinkle or buckle under large deformation, resulting in inconsistent strain for a given load. To increase compliance, we applied a four-pointed-star pattern [16] to the surface (Fig. 2(b)), which behaves like a kirigami layout: under normal loading, the surface extends in-plane, promoting smooth indentation. However, excessive stretchability would confine deformation and shrink the receptive field; therefore, the geometry was tuned to optimize strain propagation, where the connecting part length was optimized at 2.5 mm for the 7.5 mm unit cell.

The infill layer serves to spread deformation and mechanically accommodates contact loads. We implemented it using a diamond-strut unit cell (Fig. 2(c)). This bending-dominated structure allows the struts to flex under external loads, effectively spreading deformation from localized contacts over a large area [17]. Its mechanical properties are independent of cell resolution, and the overall stiffness can be tuned predictably by adjusting the strut thickness [18].

Finally, the base layer fixes the position of the infill lattice and preserves its alignment during deformation. We implemented the base as a lattice rather than a solid surface to accommodate the selective laser sintering (SLS) process: SLS printing leaves unsintered powder within enclosed volumes, and a solid base would hinder depowdering. Accordingly, we adopted a perforated square-grid base with increased strut

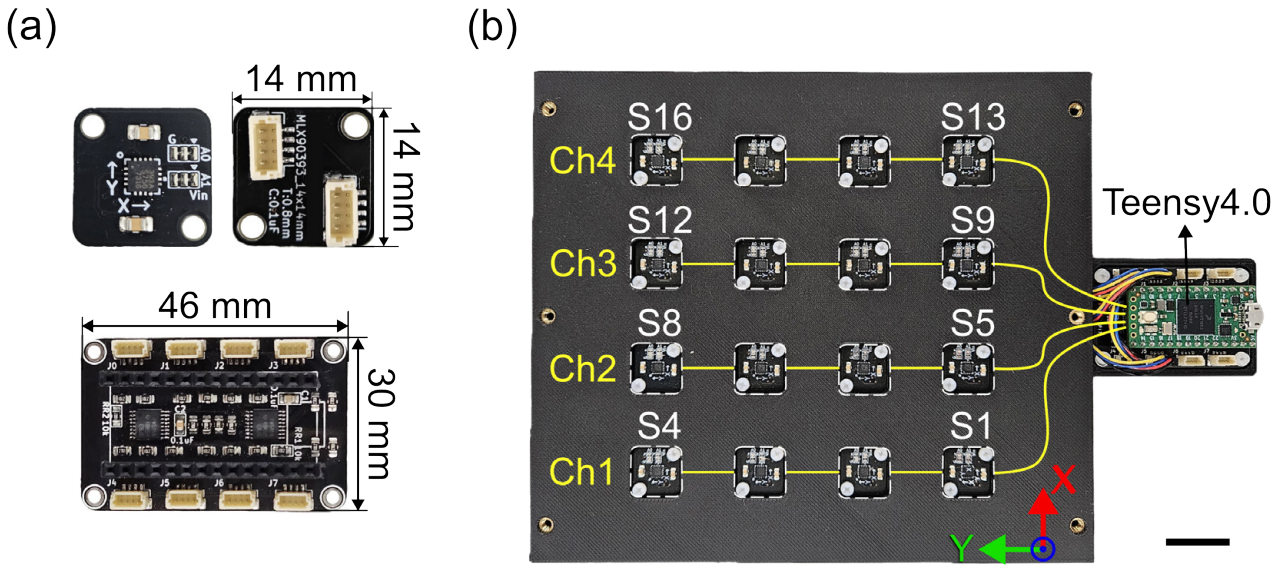

Fig. 3. Sensing electronics. (a) single Hall-effect sensor module, (b) sensor array and control board (scale bar: 2cm).

thickness to stabilize the lattice geometry and to minimize in-plane strain.

Notably, this workflow readily produces cylindrical or otherwise arbitrary geometries. Once the unit cell is specified, the implicit model automatically conforms the lattice to the target shape; this is an important advantage when designing functional, whole-body robotic skins with complex geometries.

### B. Magnetic Sensor Array

To achieve a durable and well-integrated system, we placed the sensors, wires, and microcontroller inside the robot enclosure, physically isolating them from external stimuli. Permanent magnets were embedded on the surface lattice, and magnetic hall sensor (MLX90393, Melexis) measured magnetic-field variations in a non-contact manner. We produced 14 mm × 14 mm PCBs for each Hall-effect sensor, and connected the modules over a shared serial bus. The modules interface to the microcontroller through an I$^2$C multiplexer (TCA9546A), enabling data acquisition from 16 sensors at a sampling rate of 40 Hz. Because the lattice spreads the magnetic-field sufficiently, the sensor modules could be spaced at approximately 30 mm without any blind spot.

## III. FABRICATION

The developed lattice structure was fabricated using TPU material. TPU is a flexible and elastic material that, when combined with the lattice structure, enables significant deformation capacity through its excellent elongation properties, allowing the construction of a robust layer capable of withstanding large deformations without failure. However, the structure's complex, densely packed internal features make fabrication on a common fused-deposition modeling (FDM) printer nearly impossible. By contrast, an SLS printer (Fuse 1+ 30W, Formlabs) selectively sinters a thin layer of powder (TPU 90A powder, Formlabs) with a laser, while the unsintered powder serves as a self-supporting medium. This enables straightforward fabrication of unsupported overhangs and very small features [19]. SLS also offers higher resolution and faster build speeds than FDM. Despite its size (140 mm × 140 mm × 15 mm), the multi-lattice was printed in about two hours.

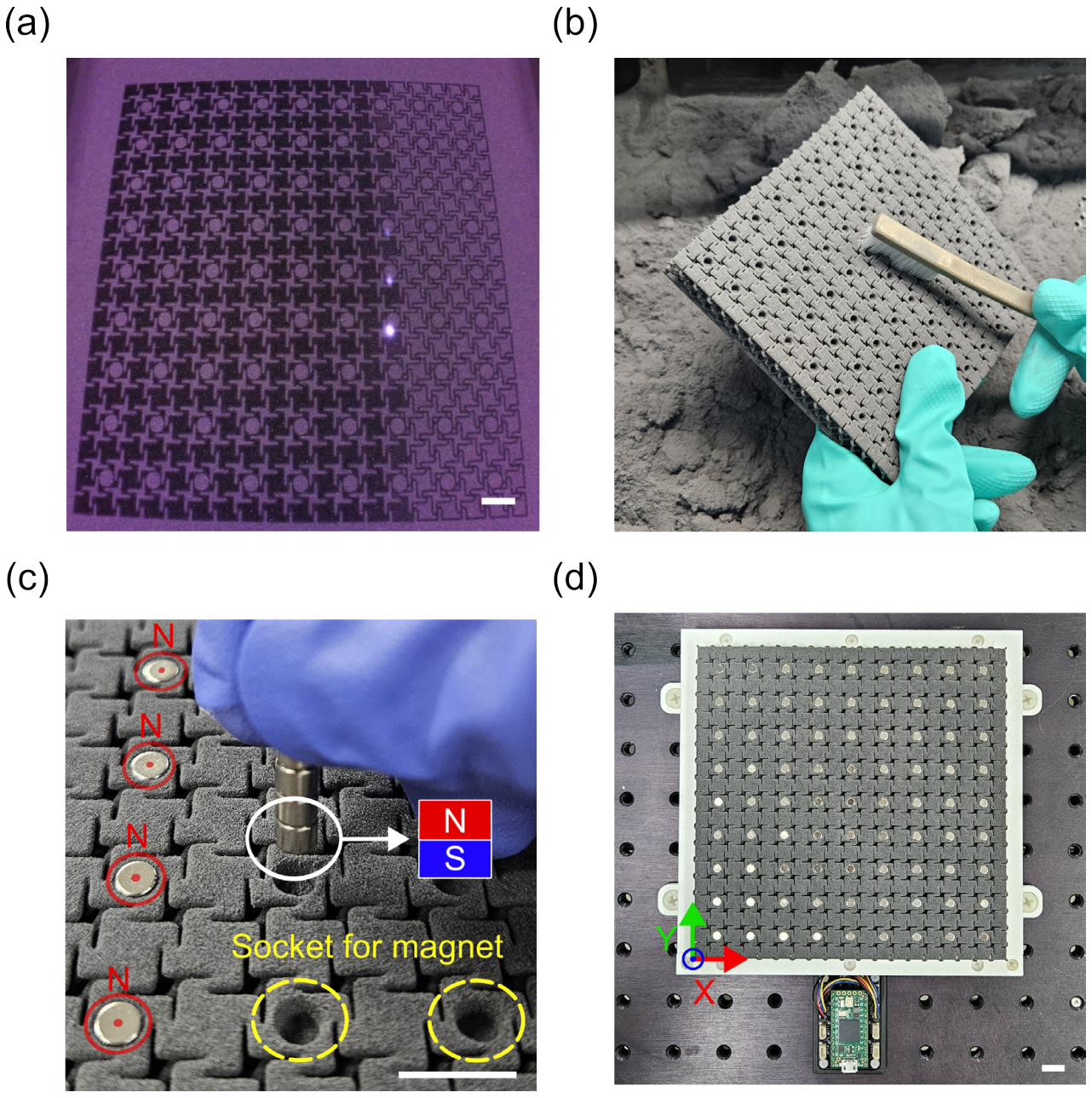

Fig. 4. Fabrication of proposed robotic skin (scale bar: 10 mm): (a) 3D printing of lattice structure using SLS printer, (b) depowdering process, (c) assembling permanent magnet, (d) fabricated robotic skin.

The overall fabrication process is shown in Fig. 4. After printing, the powder cake was allowed to cool, removed from the build chamber, and depowdered using a brush or compressed air. Miniature permanent magnets were then bonded to the surface lattice. Finally, the finished lattice was mounted on a housing with embedded Hall-effect sensors, completing the magnetic robotic skin.

## IV. TACTILE SUPER-RESOLUTION

### A. Data Collection

To map the relationship between external forces and sensor responses, we acquired data by indenting the skin with a 6-DoF robot manipulator (UR5e, Universal Robots) along a predefined trajectory. To prevent magnetic interference from nearby ferromagnetic objects, the skin was placed on a nonmagnetic metal table, and a plastic indenter tip was used. To fully characterize the sensor and avoid coverage gaps in the learning dataset, we performed out-of-plane indentations at 361 locations on a 19-by-19 grid with a 7.5 mm pitch (Fig. 5). The end-effector position was recorded at 1 ms intervals, and only samples corresponding to actual contact were retained for training. A total of 16 Hall-effect sensors mounted beneath the skin measured 3-axis magnetic fields at 41.7 Hz, resulting in 48 channels. The robot position data were synchronized with the sensor signals to construct paired input-output datasets for CNN training. For illustration, Fig. 5(c) shows time-series magnetic-field traces from 38 indentations along the y-axis (7080 mm). The panel stacks $\Delta B_x, \Delta B_y$, and $\Delta B_z$ from all 16 sensors, with time on the x-axis and channel index (sensor-axis) on the y-axis. The observed patterns reflect overlapping receptive fields of the sensors, induced by wide-area skin deformation under indentation forces. Thus, the dataset provides rich spatial features and reliable ground-truth pairs, which are essential for training the proposed CNN-based tactile super-resolution model. The current indentation experiments were conducted using a UR5e robot; however, we plan to design a dedicated linear stage in the future to achieve more precise depth and position control, thereby providing higher-fidelity training data.

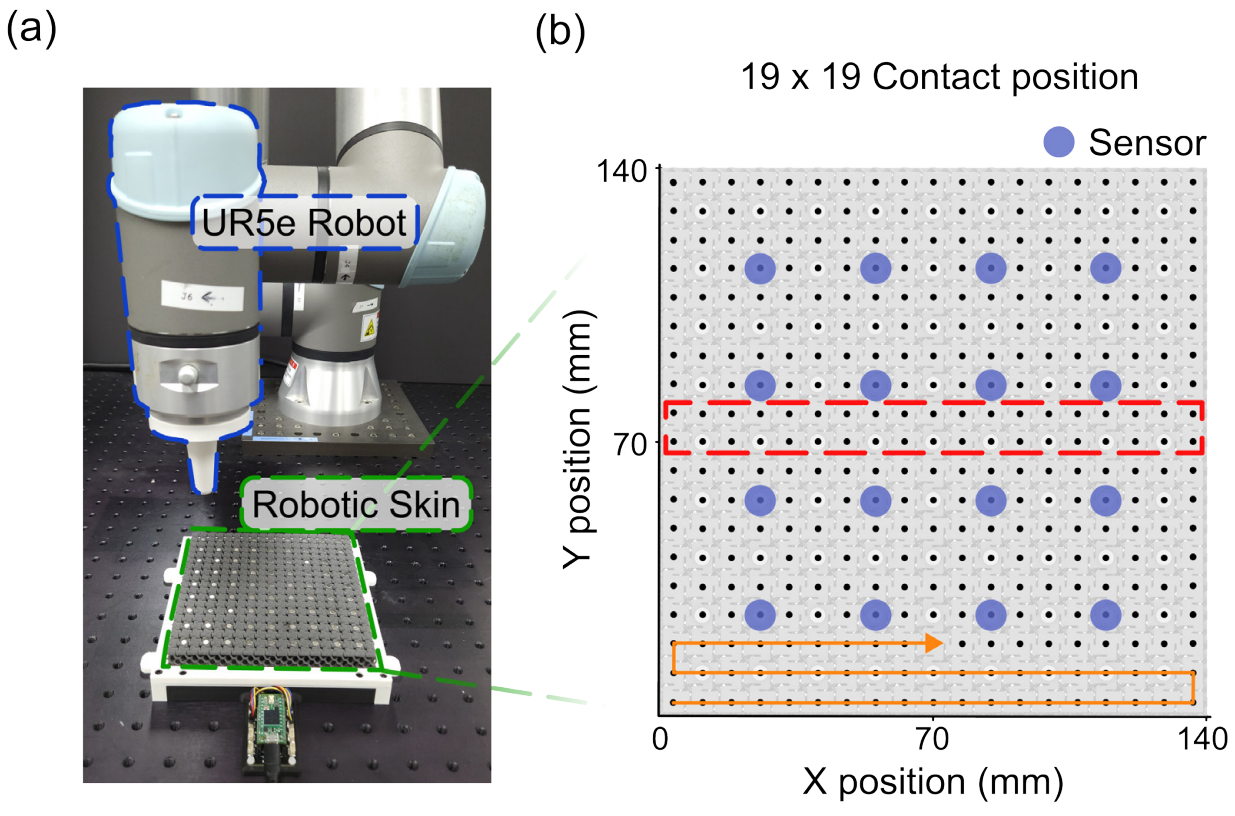

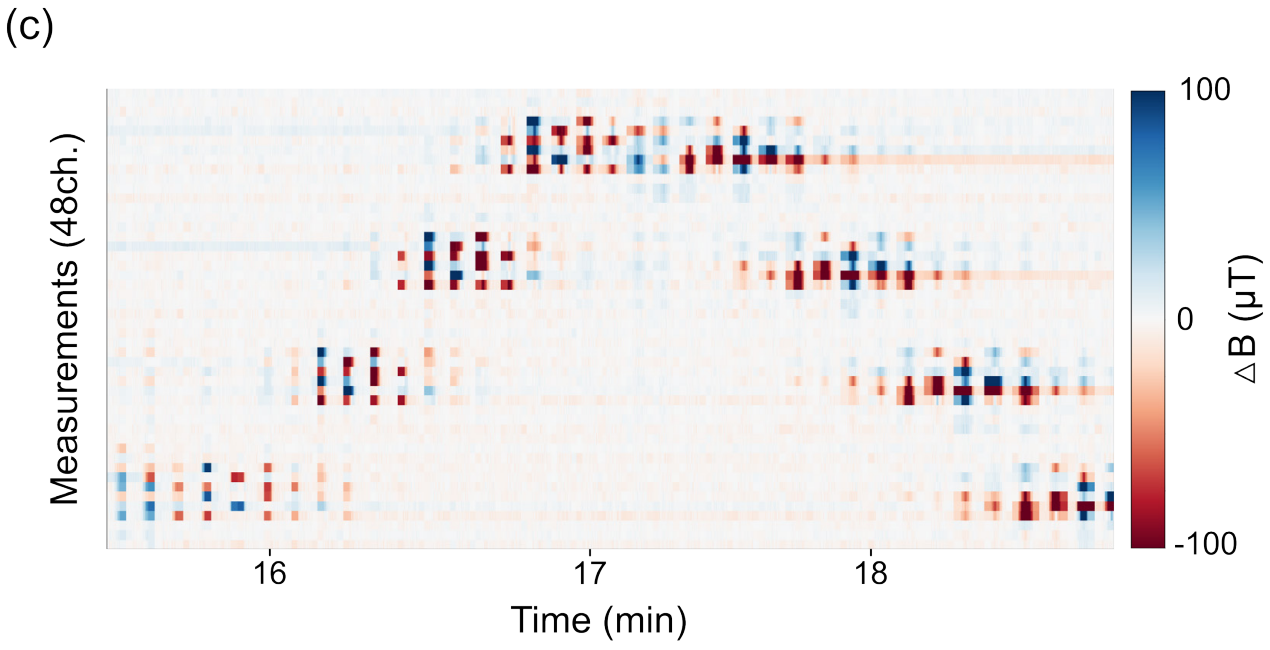

Fig. 5. Data collection process. (a) Experimental setup comprising developed robotic skin and UR5e robot; (b) Indentation was performed on 361 points along the predefined trajectory (orange); (c) Time-series plot of the 48-channel sensor data collected at the region highlighted in red.

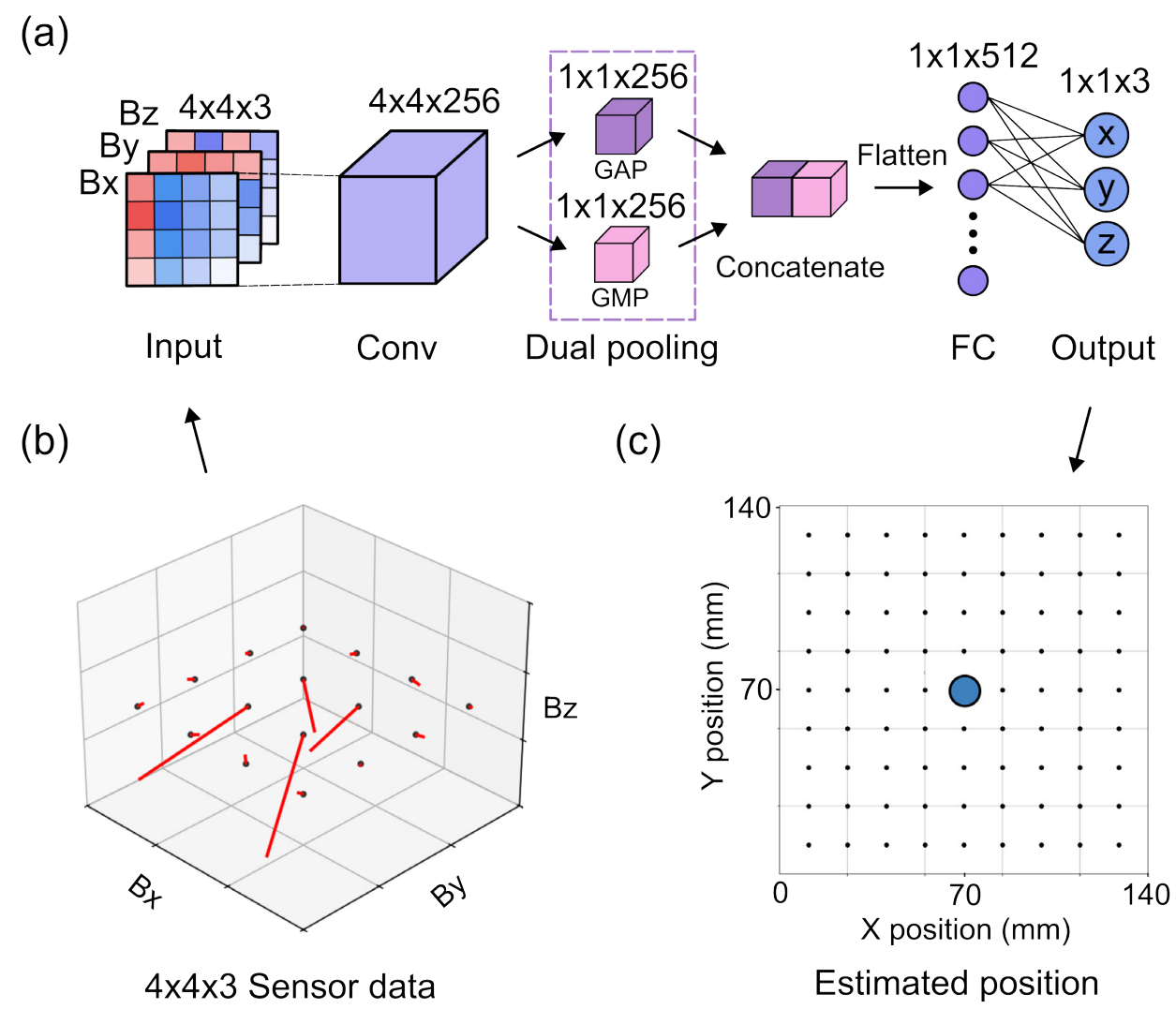

Fig. 6. Contact inference using CNN. (a) model architecture, (b) 4×4×3 sensor data vector graph (input), (c) estimated contact point (output)

### B. CNN Based Regression Model

CNN-based models have been widely used in tactile sensing to achieve super-resolution and have proven effective for feature extraction and regression tasks in sensor data [20], [21]. The proposed tactile sensing system uses a CNN to estimate contact positions from distributed magnetic sensor measurements. By learning the complex spatial relationship between sensor responses across the entire surface and actual contact positions, the CNN regression model enables super-resolution tactile sensing. As shown in Fig. 6(a), the network architecture processes 4×4×3 sensor inputs through four convolutional layers, with the number of channels progressively increasing from 32 to 256. Each convolutional layer applies a 3×3 kernel with same padding, allowing the network to capture not only local magnetic field variations but also broader spatial patterns. As a result, the initial 4×4 spatial layout is transformed into a rich 256-dimensional feature representation at each location. After feature extraction, the network leverages a dual pooling method that combines Global Average Pooling (GAP) and Global Max Pooling (GMP). GAP captures the overall magnetic response across the entire sensor array, while GMP preserves the strongest localized magnetic field responses. The combination produces a 512-dimensional feature vector that effectively encodes diverse spatial information. Finally, the fully connected (FC) layers map the learned features to three-dimensional contact coordinates. The FC stack consists of five layers (512→512→256→128→64), gradually refining the relationship between magnetic field patterns and spatial positions, before outputting continuous (x, y, z) coordinates. The network was trained using Mean Squared Error (MSE) as the loss function and the AdamW optimizer [22] with a weight decay parameter of 1e-4, which minimized the discrepancy between predicted and actual contact positions. In this way, our CNN model achieves sub-millimeter localization accuracy over a 140 mm × 140 mm sensing area (see Fig. 7), while using only 16 magnetic sensors, demonstrating efficient tactile super-resolution capability compared to conventional direct measurement methods.

## V. RESULT

### A. Contact Localization

To evaluate the performance of the proposed tactile sensing system, we conducted validation using a separate test dataset not included in the training process. Fig. 7 visualizes the distribution of localization errors across all contact positions. Fig. 7(a) shows the XY plane errors, where the mean error was measured at 1.13 mm and the deviation remained 2.27 mm. Most regions exhibited errors within approximately 1 mm, while slightly larger deviations appeared near the edges. Fig. 7(b) presents the depth estimation errors along the Z-axis, where the average error was 0.55 mm with a deviation of 0.4 mm. Overall, the system achieved uniform and stable localization performance across the surface. This result reveals that the integration of the multilayer lattice structure

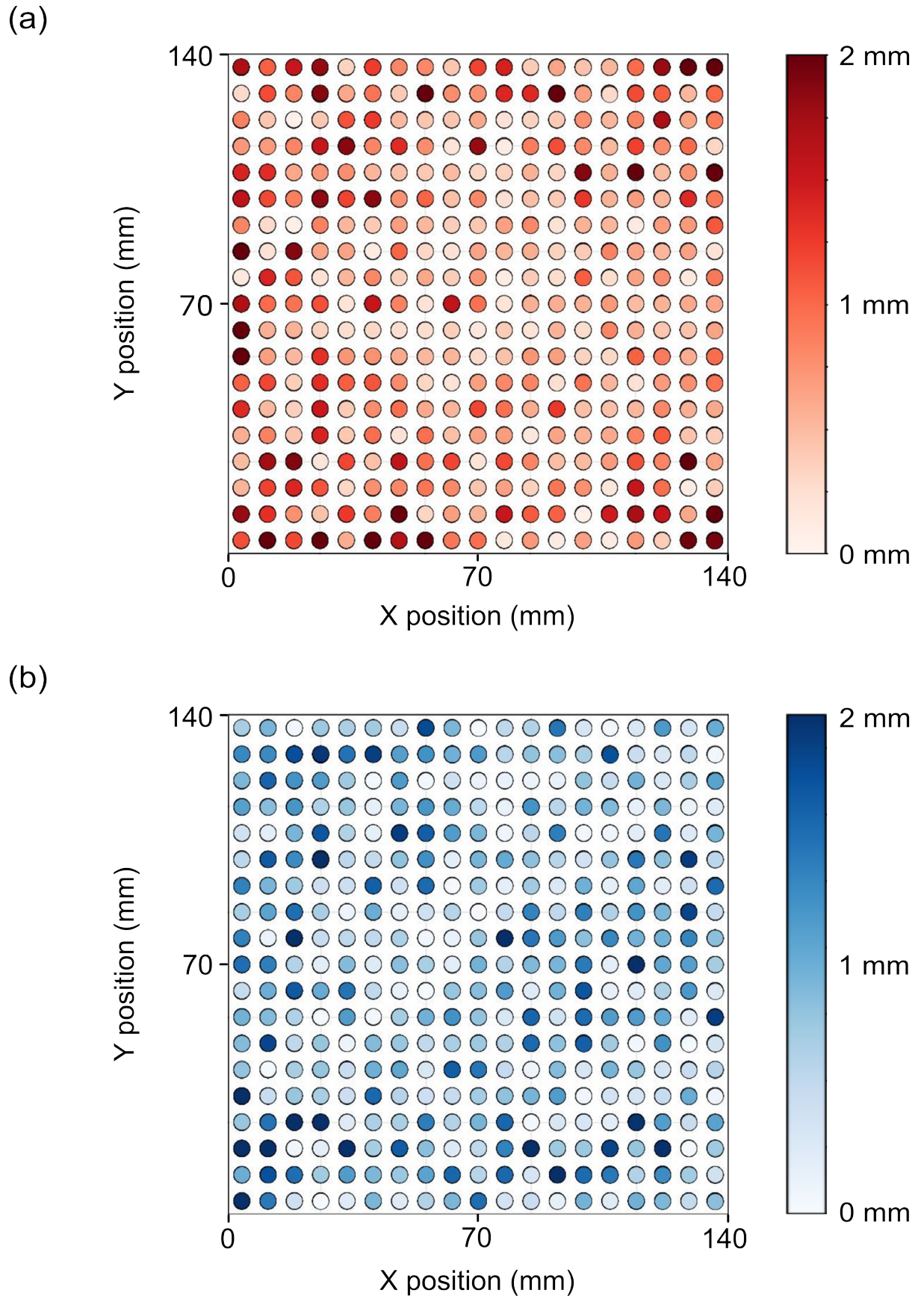

Fig. 7. Contact localization and indentation depth estimation. The color of each circle indicates the magnitude of the estimation error at that position. (a) XY position (mean: 1.13mm, deviation: 2.27mm); (b) Z position (mean: 0.55mm, deviation: 0.4mm)

with the CNN model enables precise contact localization over a wide area using only a small number of sensors.

### B. Real-Time Inference

Fig. 8 presents the real-time contact estimation results obtained using the proposed robotic skin integrated with the trained CNN model. When a fingertip pressed the skin surface, the system accurately calculated the contact position and depth with sub-millimeter accuracy and visualized the results on a 2D graph. Fig. 8(a) shows the case of pressing at the center of the skin, while Fig. 8(b) illustrates the case at the upper-right corner, both demonstrating stable and consistent tracking performance across different contact locations. These results validate that the proposed robotic skin can immediately detect and respond to external contacts through a data-driven model without requiring complex computations, highlighting its potential for real-time applications in humanoid and collaborative robots. Further details and a demonstration of the system can be found in the supplementary video S1.

## VI. DISCUSSION

This work introduced a magnet-based robotic skin that leverages tactile super-resolution. The core idea is a soft,

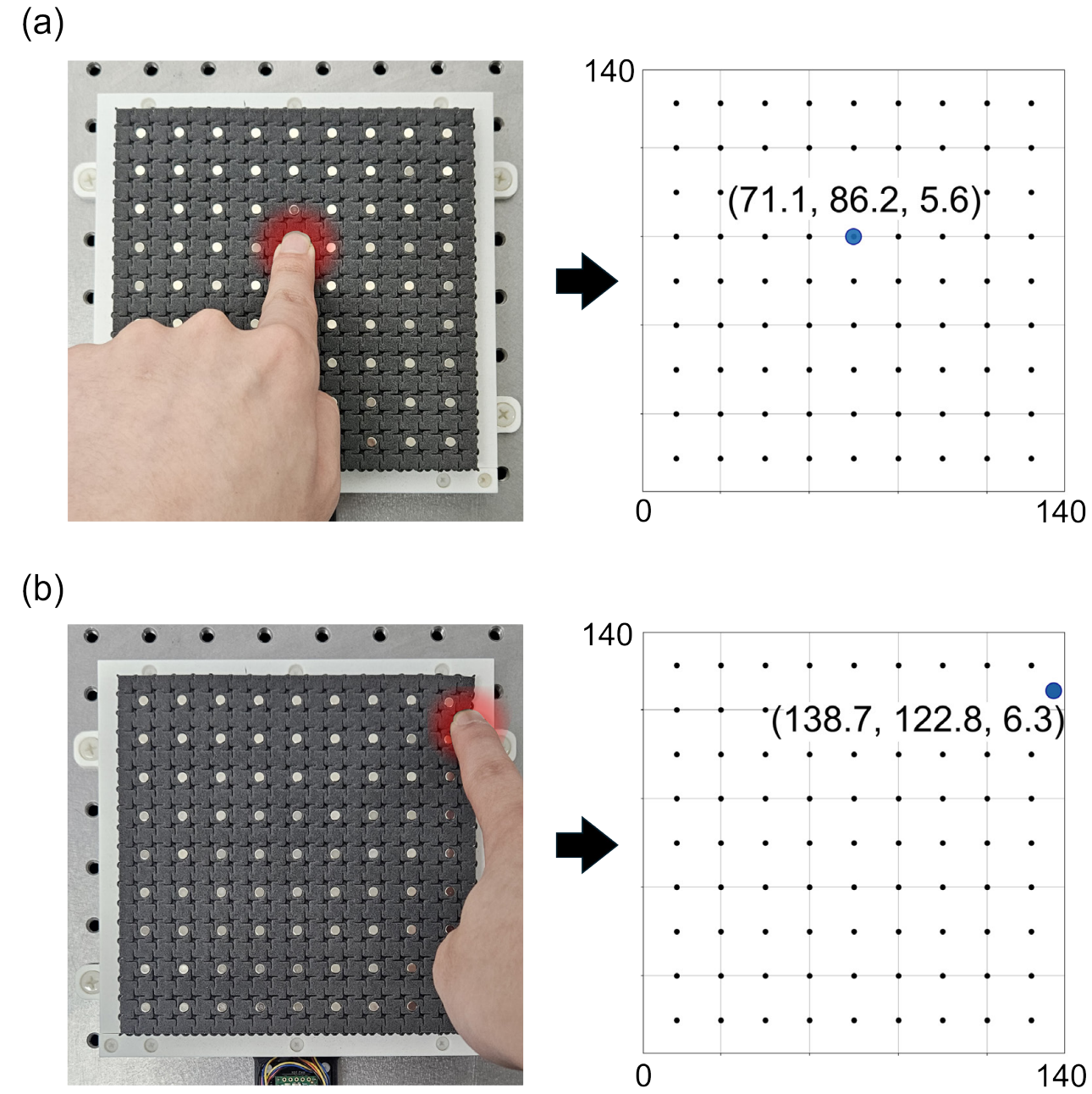

Fig. 8. Real-time contact position estimation. The developed robotic skin is indented at (a) the center and (b) the corner.

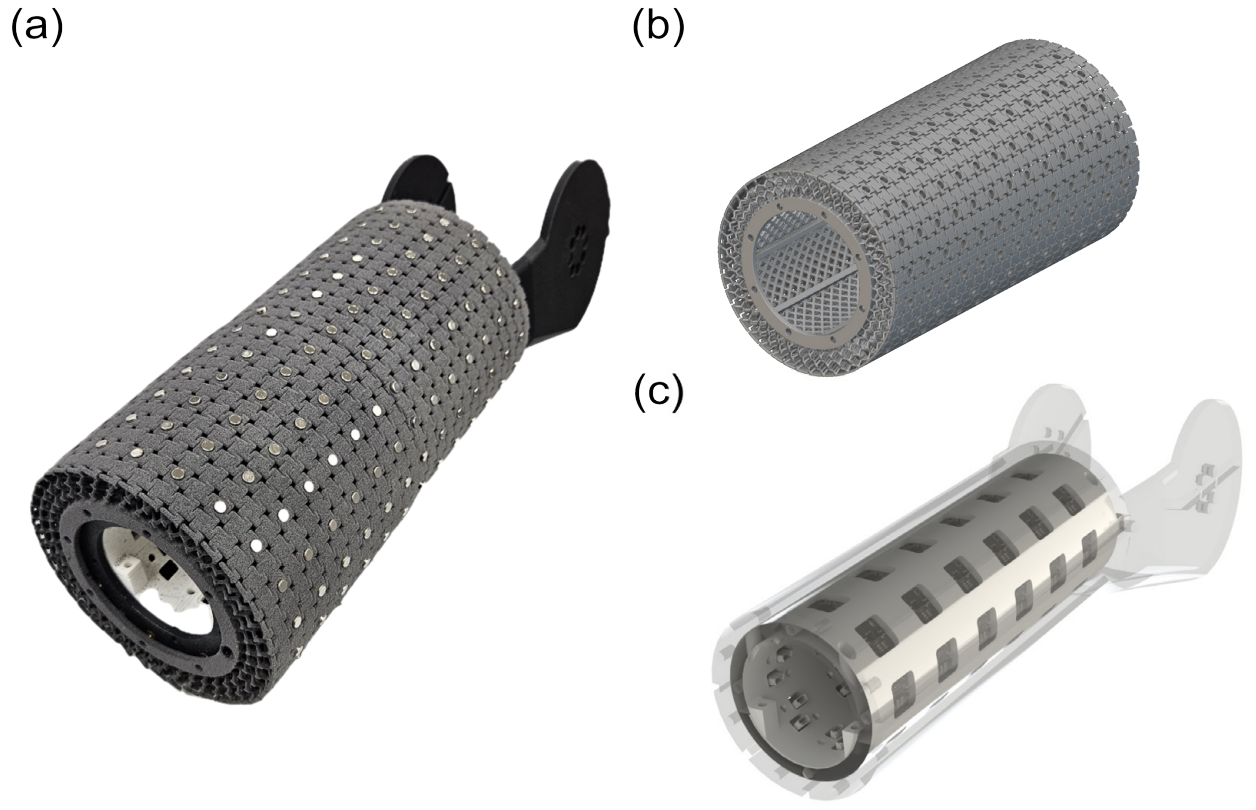

Fig. 9. Demonstration of large-scale implementation: (a) demonstration on a humanoid forearm, (b) lattice layer and (c) sensor array design on curved surface

lattice structure that converts external forces into perturbations of the magnetic field and intentionally spreads those perturbations from the contact site. By shaping the spatial spread, each sensor attains a large, overlapping receptive field, enabling sparse sensor layouts while maintaining wide-area coverage. Moreover, we could tune the lattice parameters to optimize mechanical compliance and transduction characteristics at the same time.

From a manufacturing standpoint, the lattice was generated via an implicit modeling workflow and fabricated using SLS 3D printing, which supports rapid iteration and conformal, high-complexity geometries. We demonstrated a cylindrical skin that mounts on a robot forearm, suggesting scalability to whole-body tactile coverage (Fig. 9).

We also showed that tactile super-resolution is feasible in this setting. Because a single contact stimulates multiple sensors, the resulting signal redundancy enables reconstruction at spatial resolutions exceeding the physical sensor density. Training on real-world data improved robustness to modeling errors often seen in sim-to-real transfer. Looking forward, simulation-based augmentation could cover rare or hard-to-measure stimuli, further expanding the operating envelope.

Future work will target multi-axial force sensing to estimate shear and normal forces. We also plan simulation-based parameter studies to quantify the relationship between lattice geometry, stiffness, and receptive field extent, and closed-loop integration with robot control systems for safe physical human-robot interaction.

## ACKNOWLEDGMENT

This work was supported in part by the National Research Foundation of Korea (NRF) grant funded by the Korea government (MSIT) (RS-2024-00352818), in part by the NRF grant funded by the MSIT (RS-2025-25448259), in part by Basic Science Research Program through the NRF funded by the Ministry of Education (RS-2025-25420118), and in part by the Institute of Information & Communications Technology Planning & Evaluation (IITP) grant funded by the Korea government (MSIT) (RS-2025-25442149, LG AI STAR Talent Development Program for Leading Large-Scale Generative AI Models in the Physical AI Domain).